\documentclass[10pt,twocolumn,letterpaper,shortlabels]{article}

\usepackage{cvpr}              

\usepackage[dvipsnames]{xcolor}

\usepackage[utf8]{inputenc} 
\usepackage[T1]{fontenc}    
\usepackage{url}            
\usepackage{booktabs}       
\usepackage{amsfonts}       
\usepackage{nicefrac}       
\usepackage{microtype}      
\usepackage{xcolor}         
\usepackage{amsmath}         
\newcommand{\approach}{\text{\textit{RadarSim}}}
\usepackage{graphicx}

\usepackage[dvipsnames]{xcolor}
\usepackage{amsmath}
\usepackage{amsthm}
\usepackage{amsfonts}
\usepackage{amssymb}
\usepackage{graphicx}
\usepackage{adjustbox}
\usepackage{multicol}
\usepackage{mathtools}

\usepackage{bm}
\usepackage{soul}
\usepackage{caption}
\usepackage{subcaption}
\usepackage{multirow}
\usepackage{tikz}
\usepackage{float}
\usepackage{wrapfig}

\definecolor{originblue}{RGB}{68,114,196}

\definecolor{cvprblue}{rgb}{0.21,0.49,0.74}
\usepackage[pagebackref,breaklinks,colorlinks,citecolor=cvprblue]{hyperref}

\def\paperID{406} 
\def\confName{3DV\xspace}
\def\confYear{2026\xspace}

\title{RadarSim: Simulating Single-Chip Radar via
Multimodal Neural Fields}

\author{
Chuhan Chen$^1$
\and
Tianshu Huang$^{12}$
\and
Akarsh Prabhakara$^3$
\and
Chaithanya Kumar Mummadi$^2$
\and
Zhongxiao Cong$^1$
\and
Anthony Rowe$^{12}$
\and
Matthew O'Toole$^1$
\and
Deva Ramanan$^1$
\\ 
\and
$^1$Carnegie Mellon University
\and
$^2$Bosch Research
\and
$^3$University of Wisconsin–Madison 
\and
{\tt\small \url{sally-chen.github.io/radar-sim}}
}

\begin{document}

 \twocolumn[{
 \begin{center}
 \maketitle
 \includegraphics[width=\textwidth,clip=true,trim=0mm 0mm 0mm 0mm
 ]{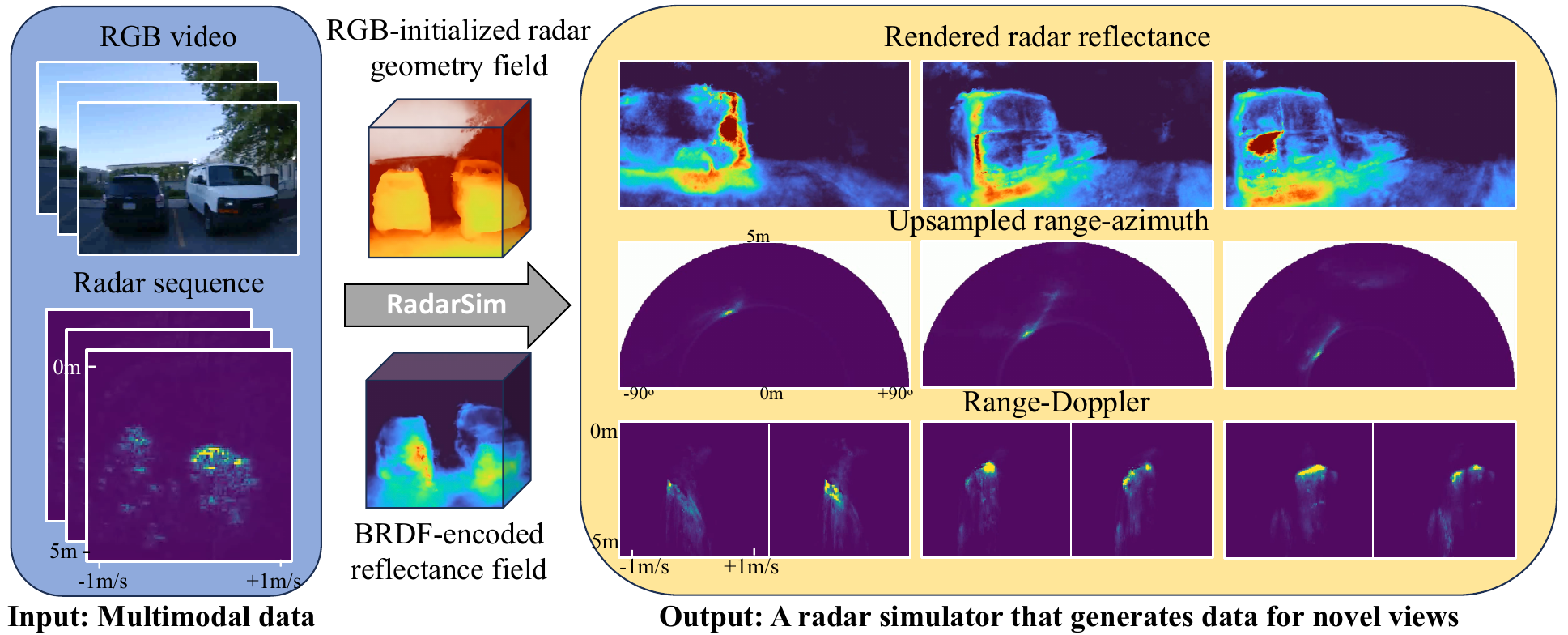}
     \captionof{figure}{Given synchronized measurements from a mmWave radar and RGB camera, we learn a spectral field model for rendering “superresolution” radar reflectance \textbf{(top right)}, which has higher fidelity and interpretability than the input radar sequence \textbf{(bottom left)}.}
 \label{fig:teaser}
 \end{center}
 }]

\begin{abstract}
Radars are an ideal complement to cameras: both are inexpensive, solid-state sensors, with cameras offering fine angular resolution, while radars provide metric depth and robustness under adverse weather. However, radar data is 
more difficult to interpret than camera images and varies significantly between sensors, necessitating increased reliance on simulation for prototyping sensors and processing pipelines. Recent work treating radar reconstruction as a novel view synthesis problem has shown great promise in reconstructing radar-relevant geometry and simulating low-level radar data. However, such methods are constrained by the low spatial resolution of the underlying radar. To address this, we propose a unified differentiable renderer, \approach, which leverages the high angular resolution of RGB cameras to generate Doppler radar range images from a camera-initialized neural field. Using a novel data set of calibrated radar camera recordings from a custom hand-held rig, we demonstrate that~\approach\ produces sharper geometry and Doppler range frames than radar-only reconstructions.
\end{abstract}

\section{Introduction}

\label{sec:intro}

Low-resolution single-chip mmWave radars are widely used in driver assistance~\citep{rasshofer2005automotive,waldschmidt2021automotive}, collision avoidance~\citep{jansson2005collision}, agriculture~\citep{steele2017radar}, and smart homes~\citep{wan2014gesture} due to their low cost, robustness, and ability to measure absolute range. However, unlike camera images or Lidar depth maps, raw range-Doppler radar data are difficult to interpret in 3D due to angular ambiguity and cross-bin effects like side lobes and bleed \citep{li2021signal}. Additionally, radar data are heavily sensor-dependent, making it impractical to test or train radar processing algorithms on generic datasets, unlike camera-based models using internet-sourced images. 
To address these challenges, many radar reconstruction techniques extract radar-relevant geometry from multiple radar scans \citep{laviada2017multiview, laviada2017multistatic,laviada2018multiview,crosetto2000radargrammetry}, while radar simulations \citep{auer2016raysar,hirsenkorn2017ray,schoffmann2021virtual,schusler2021realistic} help simulate new sensors, test algorithms, and augment datasets \citep{bialer2024radsimreal}.

Vision-as-inverse graphics approaches inspired by Neural Radiance Fields \citep{mildenhall2021nerf} have shown great promise as data simulators. These approaches \citep{huang2024dart,borts2024radar} unify 3D reconstruction and simulation as a novel view synthesis problem, and can accurately recover high-resolution radar geometry and simulate radar scans. 

However, these methods rely solely on radar data, which has inherently low spatial resolution. This limitation prevents the capture of fine geometric details, leading to blurred reconstructions and a loss of intricate features that cameras or LiDAR can easily capture,

thereby limiting their suitability for high-fidelity novel view synthesis. 
As a result, while radar-relevant geometry can be recovered, overall quality remains inferior to the state-of-the-art camera and LiDAR-based techniques ~\citep{mildenhall2021nerf, huang2023neural}.

To bridge the gap between radar and camera-based reconstruction, we propose \approach, a unified differentiable renderer that combines radar's depth sensing with the high spatial resolution of cameras. It employs a differentiable multimodal scene representation to generate mmWave range Doppler frames with geometry initialized from a pretrained RGB neural field for enhanced detail. 

\vspace{-10pt}\paragraph{Key challenges.} While mmWave radars and RGB cameras largely share the same underlying spatial geometry, their properties can differ significantly. Radars process electromagnetic spectra at millimeter wavelengths, while visible light consists of spectra at nanometer wavelengths. This can cause dramatic differences in wave propagation across space and wave reflection at surfaces.

For instance, mmWave radars perceive glass as opaque but see plastic bodywork and thin walls as transparent. 

Thus, radar-camera reconstruction must align their shared geometry while allowing for modality-specific differences.

Our key insight is that radar field geometry can be initialized and regularized with camera-field geometry (learned from a pre-trained RGB neural field). This allows our approach to preserve fine details provided by camera while accurately simulating radar measurements, ensuring high-resolution reconstruction with radar-consistent depth. 

\begin{figure*}[t]
\centering
\includegraphics[width=\textwidth
]{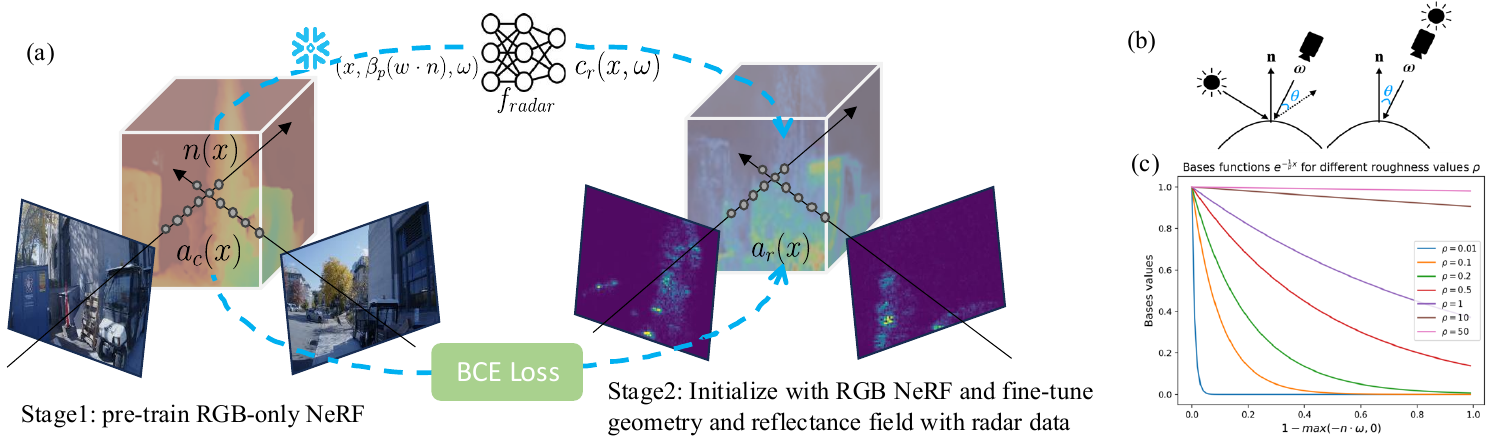}
    \vspace{-1.5em}
\caption{\textbf{(a)} 
\textit{RadarSim} uses volumetric radar reflectance and occupancy models to render a high resolution radar reflectance image. Importantly, it initializes (and regularizes) the radar occupancy model to be similar to a pre-trained RGB-NeRF occupancy model. 
\textbf{(b) left: } To better model radar reflectance, we repurpose classic specular reflectance models (e.g. Phong shading~\cite{forsyth2003modern}), where the strength of the viewed specularity depends on the angle between the viewing direction and the reflected light source (reflected about the surface normal). \textbf{ (b) right:} Radars make use of co-located transmitters and receivers, implying the brightness of specular {\em retro-reflectors} will be determined by the angle between the viewing direction and surface normal. \textbf{(c)} We show BRDF basis functions used to capture the degree of view-dependant retroreflectance given by a "roughness" value $\rho$; for large $\rho$, there is little view-dependance, implying the surface has the same radar reflectance regardless of viewing angle.}
\vspace{-7pt}
\label{fig:arch}
\end{figure*}

Moreoever, surfaces tend to appear more specular under the large wavelengths of radar, which is often manifested as view-dependant {\em retro-reflection} (Fig.~\ref{fig:arch}). This provides an additional opportunity for information sharing: by representing the radar's view-dependence using a Bidirectional Reflectance Distribution Function (BRDF) relative to a (learned) surface normal, \approach\ can more accurately represent retro-reflective surfaces. 

\vspace{-10pt}\paragraph{Contributions.} We propose \approach, the first multimodal neural field to combine radar with RGB modality in a unified framework. Our contributions are as follows:
\begin{enumerate}[(1), noitemsep]
  
\item We introduce a camera-radar based framework that leverages camera geometry as a prior to learn radar-specific geometric properties (Sec.~\ref{sec:geom}). We also propose a camera-initialized proposal network for radar ray-tracing which allows \approach\ to focus on surfaces and correctly model radar's ability to see through some camera-opaque materials (Sec.~\ref{sec:sampling}).
\item Unlike prior work, we also model radar's specular retroreflectance using a novel BRDF-based encoding with learned surface normals, which provides further information sharing between camera and radar geometry (Sec.~\ref{sec:brdf}). 

\item We introduce a LiDAR-free metric scale optimization method that refines scale-less COLMAP-derived camera poses by leveraging structural cues from radar’s range-Doppler data, ensuring accurate multimodal alignment and eliminating the need for expensive LiDAR calibration.
\item Finally, due to a lack of multimodal camera-radar datasets catered toward low-cost (i.e., low-resolution) radars and multiview settings (see Tab. \ref{tab:other_datasets}), we introduce a new radar-camera dataset and demonstrate that our multimodal architecture improves radar novel view synthesis both qualitatively and quantitatively, while also enhancing density estimation of occluded surfaces (Sec.~\ref{sec:experiments}).
\end{enumerate}

\begin{table}
\centering
\footnotesize
\begin{tabular}{lllll}
\toprule
Dataset & Radar Type & Raw Data & Varying View Dir. \\
\midrule
RadarSim (Ours) & \textbf{Low Res} & \textbf{Yes} & \textbf{Yes}  \\
RADDet \cite{zhang2021raddet} & \textbf{Low Res} & \textbf{Yes} & No   \\
RADIal \cite{rebut2022raw} & High Res & \textbf{Yes} & No \\
K-radar \cite{paek2022k} & High Res & \textbf{Yes} & No \\
Coloradar \cite{kramer2022coloradar} & High Res & \textbf{Yes} & \textbf{Yes}  \\
RaDICal \cite{lim2021radical} & \textbf{Low Res} & \textbf{Yes} & No  \\
\bottomrule
\end{tabular}\vspace{-5pt}

\caption{\textbf{Comparison with other RGB + mmWave radar datasets with raw data.} We capture a dataset using a low-resolution single-chip radar and cover scene content from multiple views directions and positions.} 
\label{tab:other_datasets}\vspace{-10pt}
\end{table}

 \section{Related Works}
\label{sec:related_work}

\vspace{-5pt}\paragraph{Data-driven radar simulation.}
While there exist model-based simulators \cite{coleman1998ray,auer2016raysar,hirsenkorn2017ray,schoffmann2021virtual,schusler2021realistic,malinen2013elmer,di2004migration,furse1990improvements,teixeira1998finite} that simulate radar signals based on a known environment, we focus on data-driven radar simulation methods that infer environments from real radar measurements.   Sparse methods detect individual reflectors using CFAR (constant false alarm rate) techniques \cite{rohling1983radar, minkler1990cfar, doer2021yaw}. In contrast, dense methods represent the environment as a voxel grid, estimating radar properties for each cell. These dense methods can be either coherent (e.g., using Synthetic Aperture Radar \cite{mamandipoor201460,yamada2017high,yanik2019near,prabhakara2020osprey,qian2020,Mostajabi_2020_CVPR_Workshops} with precise motion or fixed paths) or incoherent \cite{laviada2017multiview, laviada2017multistatic,laviada2018multiview} (aggregating data without phase alignment). While SAR provides high resolution, it's typically unsuitable for large-scale mobile applications. Instead, incoherent aggregation—such as multi-view 3D reconstruction or radargrammetry offers a practical alternative by combining measurements from different views or sub-trajectories. Recently, deep learning based approaches such as \cite{fidelis2023generation,chen2023rfgenesis} learn generative models that simulate radar measurements from learnt radar data distributions. While fast, scalable and realistic, they are far less accurate than multi-view reconstruction based approaches which aggregate real measurement in the scene to simulate new views.

\vspace{-10pt}\paragraph{Neural fields for radar.}

While originally developed for photorealistic camera novel-view-synthesis, the neural-implicit inverse rendering approach pioneered by Neural Radiance Fields (NeRFs) \cite{mildenhall2021nerf} has also been extended to the radar domain. For example, DART \cite{huang2024dart} proposes a NeRF-like approach to simulate low-resolution mmWave radars in the range-Doppler domain using a multi-view sequence. Neural fields have also been proposed for other radar applications such as mechanical radars used in robotics and some autonomous vehicles \cite{borts2024radar} and synthetic aperture radars in in aerospace and remote sensing \cite{liu2023ranerf,ehret2024radar}.

\vspace{-10pt} \paragraph{Multimodal neural fields.} In addition to radar, NeRF-like approaches have also been applied to a wide variety of domains such as RSSI \cite{zhao2023nerf2}, imaging sonar \cite{qadri2022neural, reed2023neural} and Lidar \cite{huang2023neural}. Beyond single modalities, many have also proposed to incorporate different sensor modalities into a single neural field with conventional RGB cameras, including Lidar \cite{tao2024alignmif,herau2024soac,zhu2023multimodal}, thermal or infrared cameras \cite{hassan2024thermonerf,ozer2024exploring}, and even language embedding semantics using a camera-like rendering model \cite{kerr2023lerf,ballerini2024connecting}.

Crucially, existing NeRF+X multimodal models all seek to fuse conventional image-based NeRFs with other modalities which also share a similar ray-based rendering model. This is not the case for radar: unlike cameras (or Lidar), whose sensor model has \textit{range} ambiguity, radars trade absolute range resolution for \textit{angular} ambiguity, resulting in an orthogonal sensor model \cite{huang2024dart}.

 \section{Method}
\label{sec:method}

\begin{figure*}[t]
\centering
\includegraphics[width=\textwidth
]{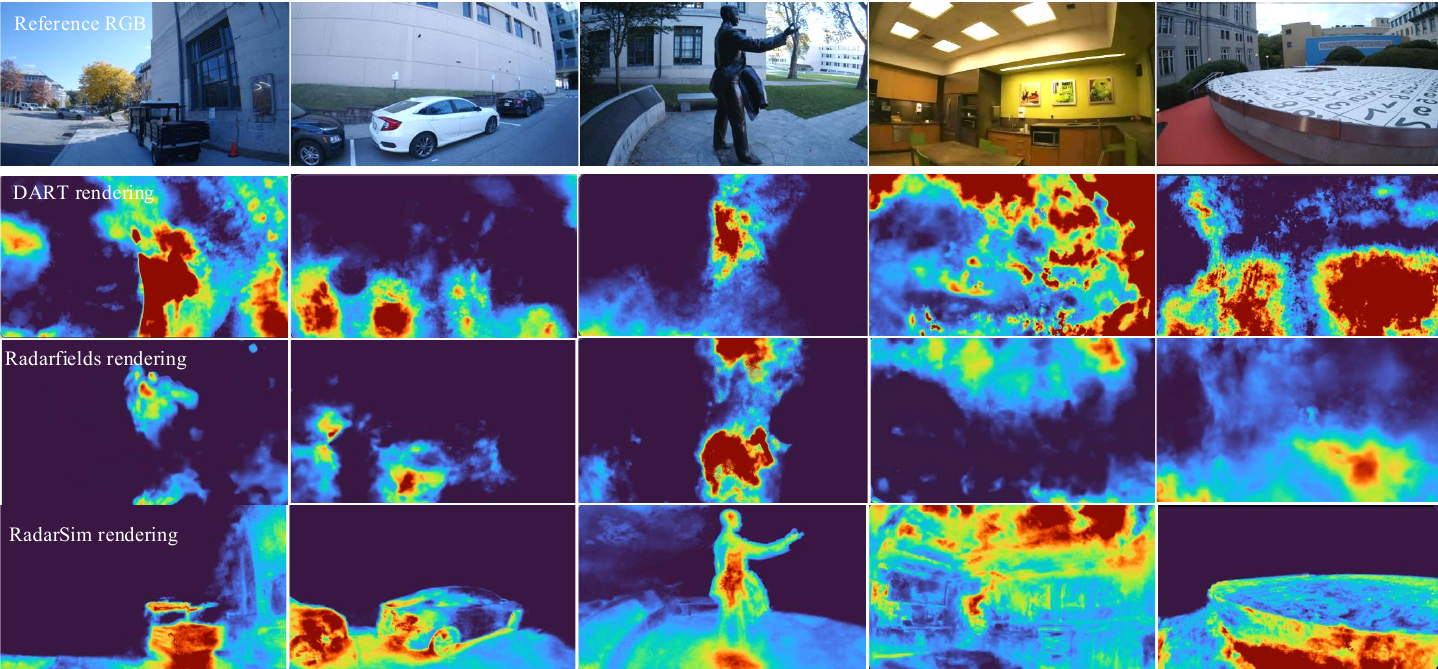}
\vspace{-1.5em}
\caption{\textbf{Novel view synthesis for DART~\cite{huang2024dart} and Radarfields~\cite{ehret2024radar} (middle) versus~\approach \ (bottom)}. Since DART and Radarfields are based solely on radar data, it is limited in spatial,   azimuth and elevation resolution and fine-detail. Specifically, they lack the ability to resolve reflectors at different heights because both doppler-range integration (DART) and range integration (Radarfields) still suffer from height ambiguity given limited data. In contrast,~\approach~combines radar with RGB camera data, and so captures sharper geometric details while faithfully recovering radar reflectance. In particular, ~\approach~models radar's characteristic {\em retro-reflectance} (Fig.~\ref{fig:arch}) as indicated by the strong responses for surfaces who's normal aligns with the camera-view (e.g., the rear of the truck), metallic surfaces, as well as concave structures like bottom of the car and corners.}
\label{fig:comparison}
\vspace{-7pt}
\end{figure*}

\begin{figure}
    \centering
    \includegraphics[width=\columnwidth]{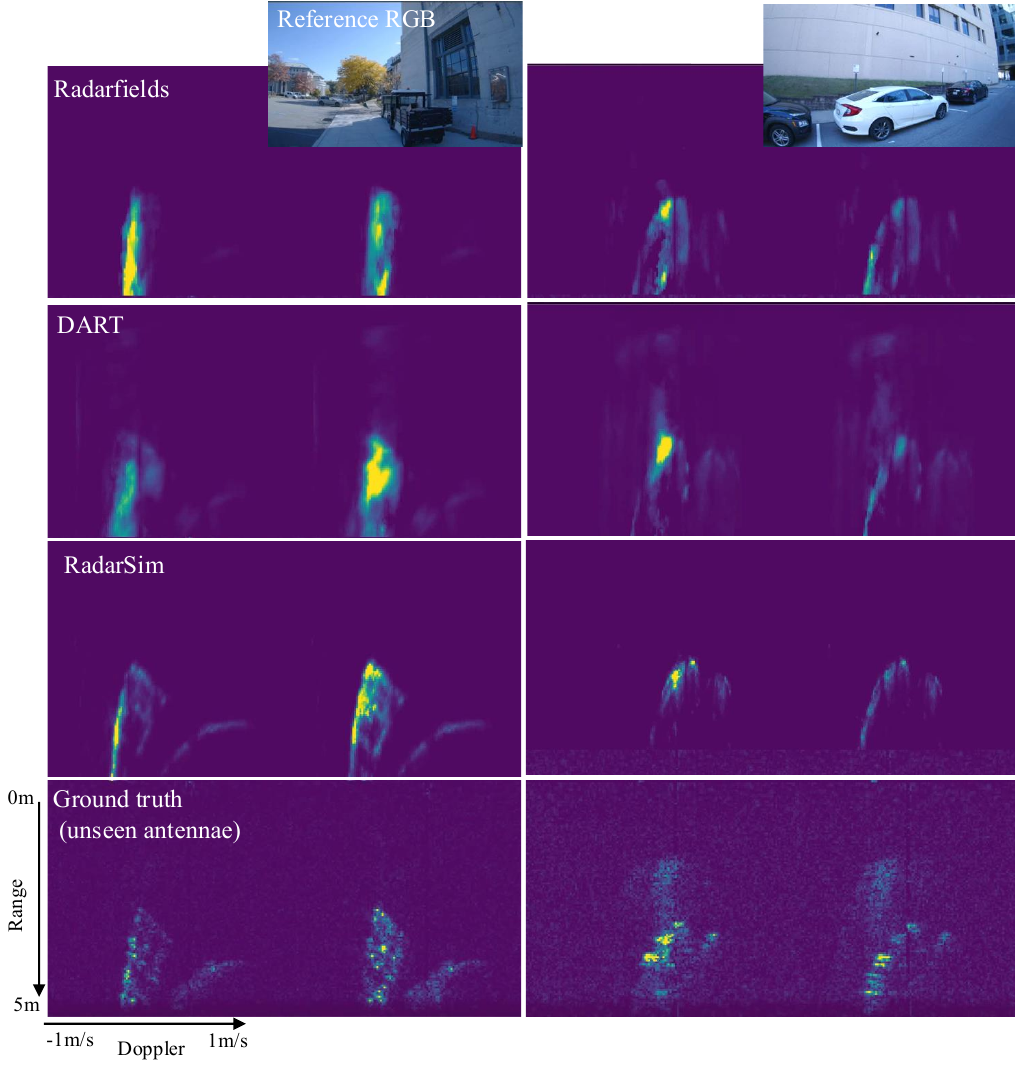}
\vspace{-1.5em}
    \caption{\textbf{Simulating unseen antennae.} \approach~can generate novel-views with modified ``intrinsics" that capture novel configurations of antennae. Here, we train \approach~on the first 4 of 8 available antennae and generate renderings of the last 4, comparing them to held-out ground-truth antennae observations. We visualize 2 of the 4 unseen antennae in this figure. \approach's renderings are sharper and closer to the ground truth. Fig.~\ref{fig:range-az} uses the same approach to generate simulations of 128 antennae, to increase the angular azimuthal resolution of the radar.}
    \label{fig:unseenant}\vspace*{-0.5cm}
\end{figure}

\approach~builds upon DART~\cite{huang2024dart}, which can be viewed as a modification of implicit neural rendering engines (NeRF \cite{mildenhall2021nerf}) for radar. Intuitively, one can view \approach~  as a unification of DART (for radar) and NERF (for RGB); given a static scene captured by synchronized radar and camera measurements, we learn a unified neural field that stores volumetric quantities that enable rendering of both RGB and radar (range-Doppler) views. However, combining both modalities is challenging. Modeling radar requires fundamentally different sampling strategies, since range Doppler ``pixel" measurements are generated by integrating along a circle in space rather than a ray (since radar waves propagate radially rather than along rays). Because of the differences in transmissive properties for radio waves and visible light, related but different geometric terms are required for camera and radar.

To capture such differences in a unified architecture, we learn a neural field for radar measurements with the help of information learnt from camera data as a regularizer. 
Specifically, from a pre-trained camera-only neural field, we initialize a geometry encoder for radar as well a proposal network for generating samples for radar ray-tracing. We model radar reflectance with an MLP conditioned on learnt camera-based geometry embedding for learning high-frequency spatially varying details. Finally, because radar tends to reflect across metallic surfaces with strong view-dependence, we model the specular reflection (that depends upon both the viewing direction and surface normal) via BRDF basis functions, repurposing techniques from implicit BRDF modeling~\cite{verbin2022ref} for capturing radar retro-reflectance.

\subsection{Background: NeRF and DART}

Since \approach~is an integration of these two frameworks, we begin by providing a unified overview of DART and NeRF; for additional details, we refer the reader to the original  references \cite{huang2024dart,mildenhall2021nerf}.

\vspace{-10pt}\paragraph{NeRF.} NeRFs learn an implicit neural field that can be used to differentiable render an image (or 2D pixel grid) by integrating volumetric {\em radiance} (or color) $\bm{c}(\bm{x}, \bm{\omega}) \in [0,1]^3$ and density $\sigma(\bm{x}) \in R$ for each 3D point $\bm{x}$ and view direction $\bm{\omega}$ along pixel-aligned ray $\bm{Y}$~\cite{mildenhall2021nerf}: 
\begin{align}
    \bm{C}(i,\bm{w}) &= \bm{c}({\bf x}_i,\bm{w}) \alpha(\bm{x}_i) \prod_{j < i} (1-\alpha(\bm{x}_j)), 
    \bm{Y} = \sum_i \bm{C}(i,\bm{w})  \label{eq:nerfC}
\end{align}
\noindent where $\alpha \in [0,1]$ are alpha-compositing weights equal to $\big( 1 - \exp{-\sigma(\bm{x}_i)\delta_i} \big)$ and $\delta_i$ is the distance between adjacent samples on a ray.

\vspace{-10pt}\paragraph{DART.} Similarly, DART learns an implicit neural field that can be used to render radar measurements, which are naturally represented as a {\em 3D cube} of range, speed (or Doppler), and angle (or antenna) measurements. To do so, DART integrates volumetric \emph{reflectance} $s(\bm{x}, \bm{\omega}) \in R$ and \emph{transmittance} $t(\bm{x}, \bm{\omega}) \in [0,1]$, capturing the proportion of energy that reflects back and that continues past a point $\bm{x}$.  These quantities can be used to model the radar return amplitude at point sample $\bm{x}_i = \bm{x} + r_i\bm{\omega}$ observed by a radar at position $\bm{x}$ with antenna $k$, written as $C(i, k, \omega)$: 
\begin{align}
    C(i, k, \bm{\omega})
     = \frac{g_k}{r_i^2} \bm{s}(\bm{x}_i,\bm{\omega})
    \prod_{j < i} t(\bm{x}_{j},\bm{w})^2,
    \label{eq:raytracing}
\end{align}

\noindent where $i$ is a discrete range bin. Compared with \eqref{eq:nerfC}, transmittance can be seen as $1 - \alpha$, but is squared since the radar signal is attenuated twice along the ray, during both the outgoing and and incoming directions after reflection. The additional inverse squared fall-off captures the radiometric reduction of energy in the reflected signal, while the antenna-dependent gain factor $g_k$ captures the dependence of the observed signal on the orientation of radar array. 

Importantly, instead of accumulating values along a ray, we must generate (or ``render") range-Doppler measurements, where the Doppler velocity of an object is its relative radial velocity. In particular, the apparent Doppler of a static point with viewing angle $\bm{w}$ captured by a moving radar with velocity $\bm{v}$ is $\langle \bm{w}, \bm{v} \rangle$: points directly in front have an apparent speed of $-||\bm{v}||$, but those off-center will have a cosine fall-off. Thus, to render a particular ``pixel" for range $r_i$ and Doppler $d_j$, we integrate samples that lie at the intersection of a {\em cone} of directions $\bm{w}$ (given a particular cosine fall-off of $d_j = \langle \bm{w}, \bm{v} \rangle$) with a {\em sphere} of radius $r_i$. Geometrically, this intersection is a circle in 3D~\cite{huang2024dart}: 
\begin{align}
    \bm{Y}(r_i, d_j, k)
    \propto \frac{r_i}{||\bm{v}||_2} \int \displaylimits_{\langle \bm{w}, \bm{v} \rangle = d_j, \ ||\bm{w}||_2 = 1}
    \mkern-36mu C(i, k, \bm{w}) \ d\bm{w}
    \label{eq:rendering-integral}
\end{align}
\noindent where the additional factors correct for the varying width of the spherical (range-Doppler) bins. 

\subsection{Sharing Geometry}
\label{sec:geom}

Given our background models, we can now define our \approach~architecture. Succinctly, we build a unified implicit neural field that generates volumetric geometry and reflectance quantities needed to render range-Doppler sensor measurements from a camera-only neural field. Two baselines (to which we compare in our ablations) are learning two separate neural fields with no sharing, as well as learning a single geometric neural field that is ``fully-shared" across camera and radar. 
The former does not allow radar to benefit from cameras, while the latter does not model the fact that geometric transmission is different across the two modalities. Instead, we first train a camera-only neural field and learn a radar geometry encoder that is regularized to be similar (but not identical) to camera geometry. 
But to do so, we reconcile an inconsistency between the two formulations: unlike NeRF (Eq.~\ref{eq:nerfC}), which fully separates geometry and radiance, DART implicitly captures scene geometry in its reflectance $s(\bm{x}_i,\bm{\omega})$ as well (Eq.~\ref{eq:raytracing}). 

Similar to other neural rendering approaches for active sensors~\cite{attal2021torf}, 
we separate reflectance into a geometry-independent reflectance term $c_r(\bm{x}, \bm{\omega})$ that captures how much energy is reflected (akin to radiance in NeRF) and a geometric-only term capturing radar-specific density $\alpha_r(\bm{x})$ that is equivalent to $1 - t(\bm{x},\bm{\omega})$. This allows us to rewrite Eq.~\ref{eq:raytracing} in a form analogous to Eq.~\ref{eq:nerfC}:

\begin{align}
    C(i, k, \bm{w})
    = \frac{g_k}{r_i^2}
    c_r(\bm{x}_i,\bm{w})\alpha_r(\bm{x}_i)
    \prod_{j < i}( 1-\alpha_r(\bm{x}_j))^2,
    \label{eq:raytracing-ours}
\end{align}   

\paragraph{Geometry encoder.} Given the modified formulation above, we now define our shared geometry encoder. We learn two neural fields for camera and radar:  
\begin{align}
   (\alpha_c(\bm{x}_i), \bm{l}_{geo_c}) = f_{geo_c} (\bm{x_i}; \theta_{geo_c}) \\[5pt]
   (\alpha_r(\bm{x}_i), \bm{l}_{geo_r}) = f_{geo_r} (\bm{x_i}; \theta_{geo_r}) 
    \label{eq:geofield}
\end{align} 
in the form of multi-resolution hash tables \cite{muller2022instant} that store geometry codes $\bm{l}_{geo_c}$ and $\bm{l}_{geo_r}$ and capture geometric properties for camera and radar respectively.  These codes are MLP-decoded into radar density $\alpha_r(\bm{x})$ and camera density $\alpha_c(\bm{x})$, respectively. The density heads are implemented as linear layers atop a shared MLP decoder. We first train $f_{geo_c}$ with camera data, and distill $f_{geo_c}$ into $f_{geo_r}$ by initializing $\theta_{geo_r}$ with $\theta_{geo_c}$. Then $\theta_{geo_c}$ is frozen and $\theta_{geo_r}$ is fine-tuned with radar data while constrained through a binary cross entropy loss between $\alpha_r(\bm{x}_i)$ and $\alpha_r(\bm{x}_i)$.

\subsection{Radar Ray Sampling} 
\label{sec:sampling}
Performance of state-of-the-art NeRF architectures such as  \cite{barron2022mipnerf360} can be attributed to efficient importance sampling on ray-surface intersections. Extending on the proposal network used in \cite{nerfstudio}\cite{barron2022mipnerf360} that generates samples from density stored in a light weight network self-supervised by the rendering weight of NeRF, we propose to share the proposal network between radar and camera and fine-tune a pre-trained proposal network for camera with rendering weight distribution of radar. While in DART~\cite{huang2024dart}, samples on radar rays are generated linearly according to range bins, we generate samples based on the sampling distribution from the proposal network, and query $f_{geo_r}$ and $f_{radar}$ to obtain $\alpha_r$ and $c_r$ for each sample on a ray. In case there are multiple samples assigned to a particular range bin, we aggregate the samples by taking the mean of the sample values; and if there are no samples, we assign 0 for $\alpha_r$ and $c_r$. We show the effect of such shared sampling scheme in geometry improvement for radar in Fig.~\ref{fig:ablate_bases} and reconstruct geometry behind occluded surface in Fig.~\ref{fig:tent}.  

\subsection{BRDF Encoding}
\label{sec:brdf}
We now describe improvements to our radar reflectance model $c_r(\bm{x}_i,\bm{w})$ that leverage improved estimates of geometry. Our motivation is that many metallic surfaces appear highly specular under radar due to its large wavelength, a phenomena sometimes known as retroreflectance. Our key insight here is to repurpose innovations from the NeRF literature on capturing surface reflectance models (BRDFs), to better model retroreflectance common in radar sensing. To do so, we augment our model to explicitly reason about surface normals and surface roughness.

\vspace{-5pt}\paragraph{Surface normals.} While normal maps could be derived by computing the spatial gradient of our geometric density model, such estimates are noisy in practice. We instead learn a MLP that predicts normals which is supervised by a monocular normal predictor on our input images~\cite{fu2024geowizard}.

\vspace{-5pt}\paragraph{Surface roughness.} Classic models of specularity compute the dot product between the viewing angle and angle of reflectance from an incident light source, where the angle of reflectance is computed by mirror-flipping the incident angle across a surface normal (Fig.~\ref{fig:arch}). However, for radars, where transmitters and receivers are collocated, the viewing and source angle are identical, implying that the quantity of interest is the dot product between the viewing angle $\bm{\omega}$ and surface normal $\bm{n}$. Retroreflective surfaces generate strong returns when viewed fronto-parallely, with a response that falls when viewed off-angle. To capture different rates of fall off, we make use of spectral basis functions 
\begin{align}
    \beta_\rho(\bm{\omega} \cdot \bm{n}) \equiv
    e^{-\frac{1}{\rho} (1-\max(-\bm{\omega}\cdot\bm{n},0))}, \quad \rho \in P.
    \label{eq:brdf-bases}
\end{align}
We now can define our final model of radar reflectance: 
\begin{align}
    \bm{c}_r(\bm{x}_i,\bm{\omega}) = f_{radar} (\bm{l_{geo_r}},  \{\beta_\rho(\bm{\omega} \cdot \bm{n})\}, \bm{\omega}; \theta_{radar}) 
    \label{eq:radarfield}
\end{align} 

\subsection{Scene Scale Optimization}

\begin{figure}
    \centering
    
    \includegraphics[width=\columnwidth]{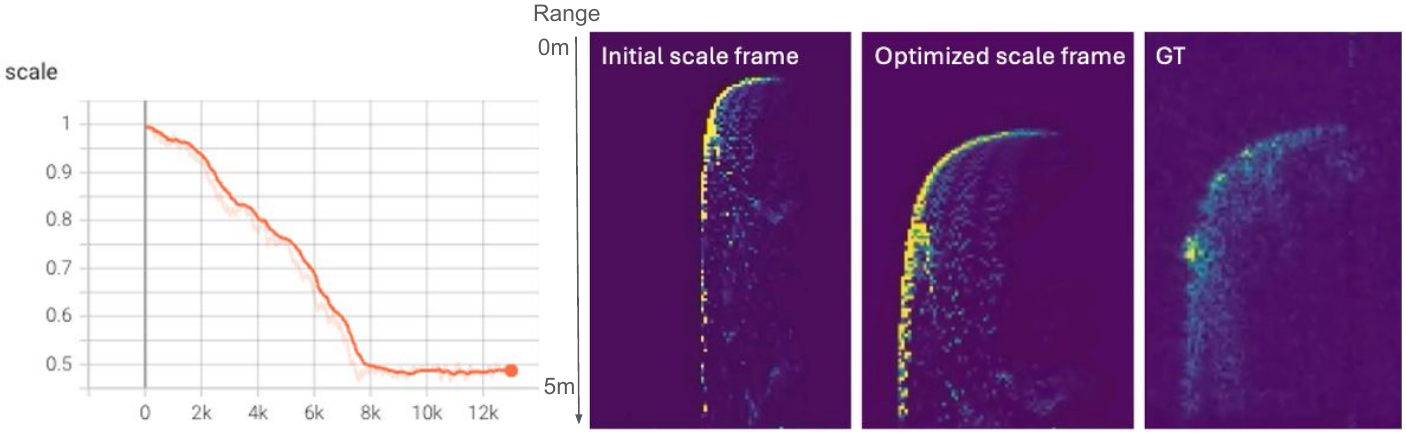}
    \vspace{-1.5em}
    \caption{\textbf{Scale optimization process.} Recall that our pipeline uses COLMAP to infer up-to-scale camera poses. We use radar to metrically-upgrade our scene reconstruction by optimizing for the scale that produces the best (metric) range-doppler reconstruction. The ({\bf left}) plot shows the optimization curve, where the scale factor adjusts from a randomly initialized value of $1$ to a final optimized value $0.483$. The ({\bf right}) images display range-Doppler renderings with the initialized scale of $1$, the optimized scale of $0.483$, and the ground truth range-Doppler frame with a scale of $0.503$. The optimized scale deviates from the ground truth by only $3.98\%$. 
}\vspace{-8pt}
    \label{fig:scale_opt}
\end{figure}

In order to optimize $f_{geo_r}$ and $f_{radar}$ using doppler integration proposed in DART~\cite{huang2024dart}, we require knowledge of the true metric sensor velocity. Because COLMAP-derived sensor poses are scaleless, it is important to obtain metric scale for each scene.  We observe that for scenes of different scales, range-doppler images contain structures that appear more expanded or compressed on the range axis as shown in Fig.~\ref{fig:scale_opt}, hence we propose to leverage our geometry sharing scheme to perform scale optimization by exploiting such structure information. From a pre-trained $f_{geo_c}$, we render range-doppler frames using $\alpha_{c}$, and use the difference in shape of the structures alone between the ground truth and synthesized radar frames to optimize for scale using only a Structural Similarity Index Measure (SSIM) loss. Refer to  Fig.~\ref{fig:scale_opt} to the optimization process and Appendix for evaluation of our learnt metric scale.

 \section{Experiments}
\label{sec:experiments}
\begin{figure*}[t]
\centering
\includegraphics[width=\textwidth
,clip=true,trim=0mm 38mm 0mm 0mm]{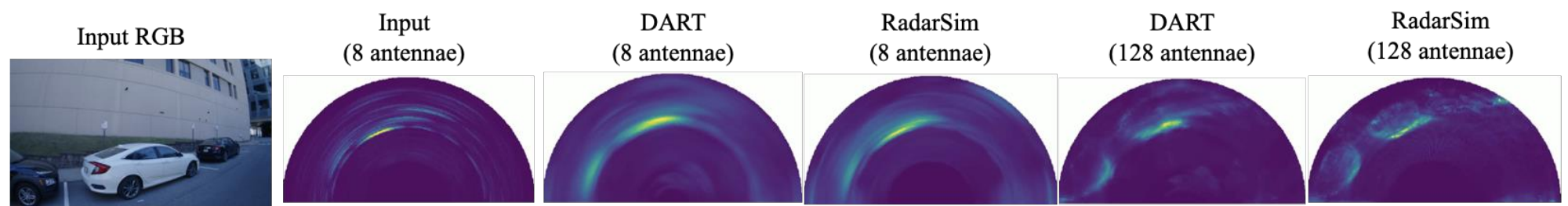}
\vspace{-1.5em}
\caption{We show range-azimuth reconstructions of ~\approach~ and DART \cite{huang2024dart}. Recall our input radar data is recorded with 8 antennae, which limits the angular azimuth resolution. 
We can use our neural reconstructions to construct plots rendered with virtual radars with any number of antennae, allowing us to "super-resolve" additional detail across azimuth angles (e.g., we can distinct parked cars with even more detail than DART).} \vspace{-5pt}  
\label{fig:range-az}    \vspace{-3pt}
\end{figure*}

We validate the efficacy of~\approach~through experiments on a diverse set of indoor and outdoor scenes.

\vspace{-5pt}\paragraph{Dataset.}
We used a handheld data collection rig 
with a radar and camera for data collection. The rig’s compact and portable design enabled us to collect 8 sequences each between 3-5 minutes duration in a diverse set of indoor and outdoor environments; we provide detailed descriptions of the data collection rig and collected traces in the Appendix.

\vspace{-5pt}\paragraph{Baselines.} 
We evaluate our multimodal framework against DART \cite{huang2024dart} and Radarfields \cite{borts2024radar}, the state of the art methods for radar-based 3D reconstruction, on the quality of reflectance and transmittance field learned by \textit{RadarSim}: rendering out reflectance by tracing rays from camera pixels into the 3D neural field and accumulating reflectance using rendering weight computed with learnt radar density.  We also evaluate on radar novel view synthesis in range-doppler frame quantitatively and qualitatively which shows the effectiveness of modelling radar reflectivity by reproducing the input signal while being capable of generalizing to unseen radar frames. We demonstrate our model can create high resolution visual rendering of radar reflectance and density, essentially a high resolution radar, from the same input low resolution radar frames. We also compare RadarSim with DART~\cite{huang2024dart} baselines (CFAR, Lidar, Nearest Neighbor) in Tab.~\ref{tab:additional-comp},  demonstrating improvements over them.

\subsection{Qualitative Results: High Resolution Radar Simulation}
We visualize the quality of \textit{RadarSim} against DART in Fig.~\ref{fig:comparison}. Our multimodal framework creates a high resolution rendering of radar reflectance field by leveraging shared geometry with RGB reconstruction, achieving a much better radar simulator than baselines. We are able to show structures that radar strongly reflects off such as retro-reflector like structures such as inset corners, bottom of cars, light on the ceiling. We also demonstrate surface normal-dependent specular reflection, e.g., when we point toward a surface, reflectance is strong. This is better viewed in our included videos in Supplemental materials. We are also able to distinguish materials such as metal, which is usually a strong reflector of radar, from walls, which are weak reflectors, as well as materials that radar signals transmit through such as glass from other non-transmissive materials. We further visualize synthesized range-Doppler frames for antennae held-out during training compared to ground truth and those synthesized by baselines in Fig. \ref{fig:unseenant}. ~\approach~'s simulation appears most similar to the ground truth, showing its superiority in generalizing to unseen view directions and radar intrinsics. We also visualize synthesized 8-antenna and simulated 128 antenna range-azimuth radar frames  in Fig. \ref{fig:range-az}, showing ~\approach~ 's ability to reconstruct sharper geometry and radar reflectance field and effectiveness in improving azimuth resolution of single-chip radar.

\subsection{Quantitative Results}
To evaluate radar novel view synthesis, we applied our model to a holdout test set consisting of the last 20\% of each sequence, and computed the Structural Similarity Index (SSIM) and Peak Signal-to-Noise Ratio (PSNR)\footnote{
    Similar to \cite{huang2024dart}, we also account for the sparsity of range-Doppler frames by ignoring regions of the range-Doppler image which are under a per-sequence noise threshold; refer to the Appendix for the procedure we used to calibrate this threshold.
} of the synthesized range-Doppler frames against their respective ground-truth frames. \approach\ achieves higher PSNR and SSIM compared to baselines (Table~\ref{tab:additional-comp}). We further evaluate view extrapolation by splitting each sequence spatially, where our model out-performs baselines by a big margin, showing our model's effectiveness in leveraging camera information for generalizable radar simulation.

\subsection{Density Estimation for Occluded Surfaces }
\begin{figure*}[t]
\centering
\includegraphics[width=\textwidth
]{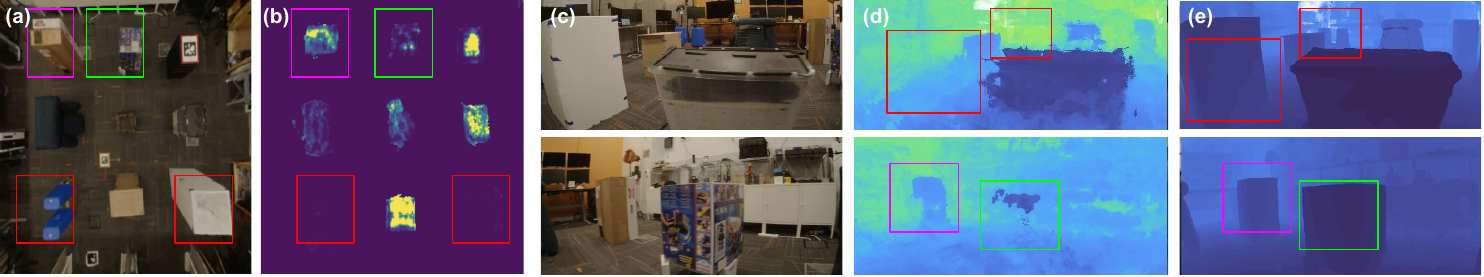}
      \vspace{-1.5em}
    \caption{RGB birds-eye view of the scene ({\bf a}), radar occupancy $\alpha_r$ slice (at 0.5m in height)  ({\bf b}) reconstructed by \approach. Reference RGB images ({\bf c}) and corresponding depth map rendering using radar occupancy  ({\bf d}) and camera occupancy ({\bf e}). Because radar transmits through  materials such as plastic cardboard, foam, etc., such geometries (annotated in {\bf red}) do not appear in the radar occupancy slices or depth renderings. We also compare boxes in purple (cardboard box with electronics) and green box (empty cardboard box), which the empty box does not show up in radar occupancy map or depth map rendered with radar occupancy.
}
    \label{fig:boxes}  \vspace*{-7pt}
\end{figure*}

\begin{figure*}[t]
\centering
\includegraphics[width=\textwidth
]{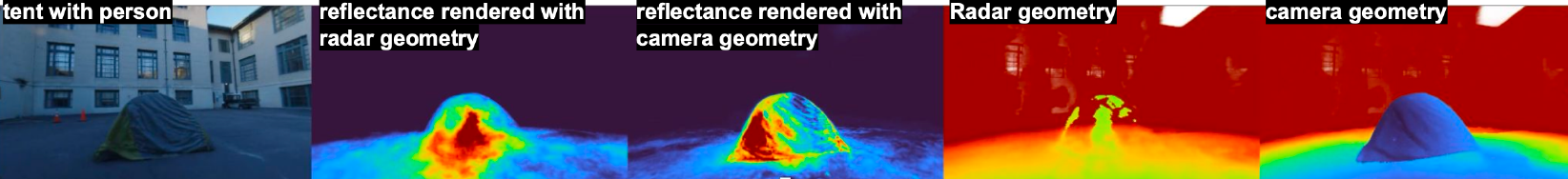}
 \vspace{-1.5em}
\caption{A tent with ({\bf top}) a person sitting inside, shown as an RGB image ({\bf left}), radar reflectance rendered from radar occupancy  ({\bf left-center}) and camera occupancy ({\bf center}), depth map rendered from radar occupancy ({\bf right-center}), and camera occupancy ({\bf right}). As radar can transmit through cloth, radar density reveals the presence of a person, while camera density is unable to do so.}
    \label{fig:tent} 
\vspace{-5pt}
\end{figure*}

Radar has the capability to penetrate certain materials that are opaque to RGB cameras, such as cloth, cardboard, and foam. Because our multimodal model estimate different geometry for radar and RGB camera,  we show in Fig.~\ref{fig:boxes} and~\ref{fig:tent} that utilizing our multimodal model, we could estimate ``emptiness" of occluded surfaces in high fidelity.

 \subsection{Ablations}
\label{sec:ablation}
\begin{figure}
    \centering
    \includegraphics[width=\columnwidth]{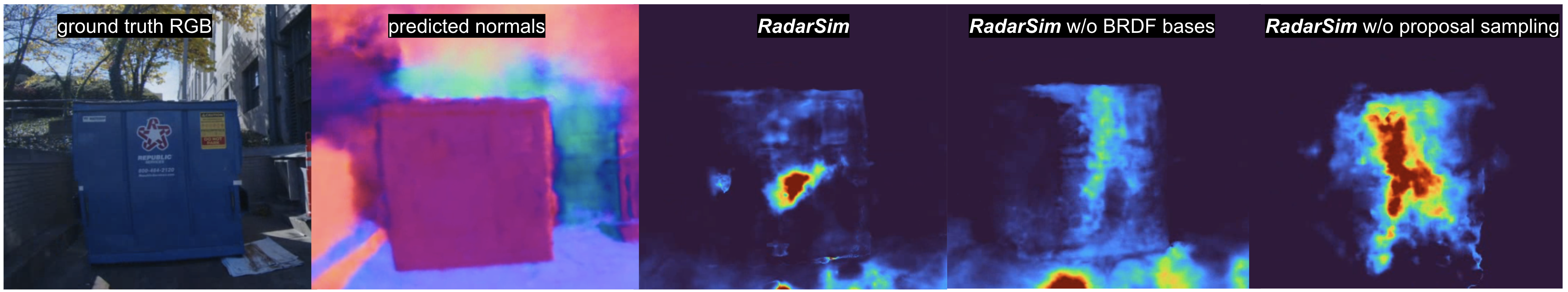}
    \caption{ Ablation on our proposed BRDF bases encoding to model view dependence (\textbf{second to the right}) and sampling (\textbf{right}). We show that we model high-frequency normal-dependent reflectance changes: when view direction points toward the surface normal, strong reflectance is shown, and quickly decreases when view-direction deviates from the normal. While conditioning with spherical harmonic encoded view direction can model such effects, the view-dependence is much lower frequency, producing a blob in place of a dot. Without proposal sampling, render radar geometry is much cloudier compared to~\approach.  } \vspace{-10pt}
    \label{fig:ablate_bases}
\end{figure}

\vspace{-5pt}\paragraph{Performance when not using BRDF bases and sampling} We analyze the effect of BRDF bases and proposal network for radar. As shown in Fig.~\ref{fig:ablate_bases}, our BRDF bases models sharp reflectance change based on angle between normal and view direction, while SH view direction encoding is much lower frequency. We also ablate on the effect of our shared proposal network with radar to generate radar ray samples; it produces sharper geometry for \textit{RadarSim}.

\vspace{-10pt}\paragraph{Ablations with ``naive" RGB-radar baselines} Tab.~\ref{tab:additional-comp} shows two additional variants of a ``naive" baseline: radar/RGB images rendered with the same occupancy predicted from a shared geometry encoder, using the same Nerfacto~\cite{tancik2023nerfstudio} and DART~\cite{huang2024dart} rendering process (excludes our geometry sharing (\cref{sec:geom}), BRDF encoding (\cref{sec:brdf}), and ray sampling (\cref{sec:sampling}) solutions.
\begin{itemize}
    \item \textbf{RGB-only geometry:} Geometry encoder is pre-trained only by RGB loss, frozen and queried by the radar MLP that learns to predict reflectance. Performance drops significantly showing it is insufficient to ``use" the geometry from RGB camera because radar has its unique transmissive properties. 
   
    \item \textbf{Fully-shared geometry:} Same as above, but geometry encoder is jointly trained using both RGB and radar losses. This remains inferior to RadarSim showing our proposed solutions effectively help reproduce radar reflectance.
\end{itemize}

\begin{table}

\footnotesize
\centering

\begin{tabular}{ccrr}
& Comparison to Prior Art & SSIM & PSNR \\
\toprule
\multicolumn{2}{c}{RadarSim} & \textbf{0.821} & \textbf{29.08} \\
\multicolumn{2}{c}{Radarfields \cite{borts2024radar}} & 0.771 & 27.80 \\
\multicolumn{2}{c}{DART \cite{huang2024dart} + pose opt} & 0.799 & 28.47 \\
\multicolumn{2}{c}{DART \cite{huang2024dart}} & 0.784 & 28.00 \\
\hline
\multirow{3}{*}{\begin{tabular}[c]{@{}c@{}}DART\\ baselines\end{tabular}} & CFAR points & 0.671 & 24.45 \\
& Lidar occupancy & 0.733 & 28.30 \\
& Nearest Neighbor & 0.725 & 25.36 \\
\toprule
\end{tabular}
\hfill
\begin{tabular}{ccrr}
& Diagnostics & SSIM & PSNR \\
\toprule

\multirow{2}{*}{\begin{tabular}[c]{@{}c@{}}Extreme\\ novel-views\end{tabular}} & DART\cite{huang2024dart} + pose opt & 0.747 & 27.14 \\ 
\multicolumn{1}{c}{} & RadarSim & 0.771 & 28.05 \\
\hline
\multirow{2}{*}{\begin{tabular}[c]{@{}c@{}} Geometry \\ ablations\end{tabular}} & RGB-only geometry & 0.771 & 28.36 \\
& Fully-shared geometry & 0.798 & 28.53 \\ \hline
\multirow{2}{*}{\begin{tabular}[c]{@{}c@{}} Architecture \\ ablations\end{tabular}} & w/o Bases & 0.802 & 28.78 \\
& w/o Sampling & 0.805 & 28.82 \\
\toprule
\end{tabular}
\vspace{-5pt}
\caption{{\bf Top:} \approach~outperforms radar-only prior art such as DART and RadarFields. We also compare to the baselines such as lidar occupancy and nearest neighbors. {\bf Bottom:} Diagnostic Analysis. We find even stronger performance deltas for extreme novel-views, by spatially splitting up scenes into a train-vs-test split (instead of splitting up scene logs by timestamp, as above). We also compare to a "naive" variant where radar occupancy is fixed to be identical to the pre-trained RGB model. Optimizing such a fully-shared model for both RGB and radar reconstruction helps somewhat, but still unperforms \approach. Finally, removing the reflectance model or the radar ray sampler modestly hurts.}

\label{tab:additional-comp}   \vspace{-10pt}
\end{table}

\section{Conclusion}

\label{sec:conclusion}

We propose \approach, which leverages the complementary properties of radar and cameras with a unified differentiable renderer to learn high-resolution radar-specific 3D geometry and simulate more accurate radar range-Doppler images. We show the applicability of~\approach~across several diverse indoor and outdoor scenes and demonstrate that implicit geometry sharing can be an incredibly powerful tool for sensor fusion in 3D reconstruction for sensors with greatly varying characteristics. While \approach\ can leverage cameras to improve radar simulation, one limitation is that the reliance on cameras for high resolution angular information may degrade performance under conditions where camera data is compromised, such as in low light or environments with shiny surfaces. Additionally, despite enhancing radar reconstructions, range-Doppler accuracy remains limited by radar's inherent spatial resolution. Motivated by the strong performance of~\approach, we believe other sensors could be integrated into a neural-implicit multi-sensor field. We envision a future system that learns shared geometry from any sensor subset and dynamically adapts to available modalities, fully leveraging sensor synergies for scalable multimodal scene understanding across diverse environments and tasks.

\section*{Acknowledgements} Chuhan Chen was supported by a NSERC Postgraduate Scholarship (PGS-D).

 {
     \small
     \bibliographystyle{ieeenat_fullname}
     \bibliography{main}
 }
\clearpage
\appendix
\setcounter{section}{0}
\renewcommand{\thesection}{\arabic{section}}

\section*{Appendix}
\addcontentsline{toc}{section}{Appendix}

\section{Supplementary Videos}
Please refer to the attached supplementary videos for results. We show comparison between \approach - Radar reflectance rendered with radar occupancy (\textbf{left column}) and  radar-only baselines DART \cite{huang2024dart}(\textbf{middle column}) and Radarfields\cite{borts2024radar}(\textbf{right column}). Observe that compared to the baseline, our reconstruction quality shown in depth map at the bottom is significantly better, while being able to accurately model radar reflectivity. Specifically, strong reflectance is visible for radar signals:

\begin{itemize}[label=\textbullet]
    \item at inset corners that act as retro-reflectors where all transmitted rays are reflected due to bouncing at corners, such as bottom of car, inside windows, wall/ceiling intersections, light fixture on ceilings etc.
    \item where view direction aligns with surface normal 
    \item metallic material in general
\end{itemize}
We also show synthesized range-azimuth view in 128 azimuth directions, achieving super-resolution of the input radar data which only contains 8 directions from 8 antenna measurement. Notice our synthesized range-azimuth rendering is much sharper than DART\cite{huang2024dart}, while preserving radar reflectivity. 

\section{Dataset}
\begin{figure}[h!]
    \centering
\includegraphics[height=10em]{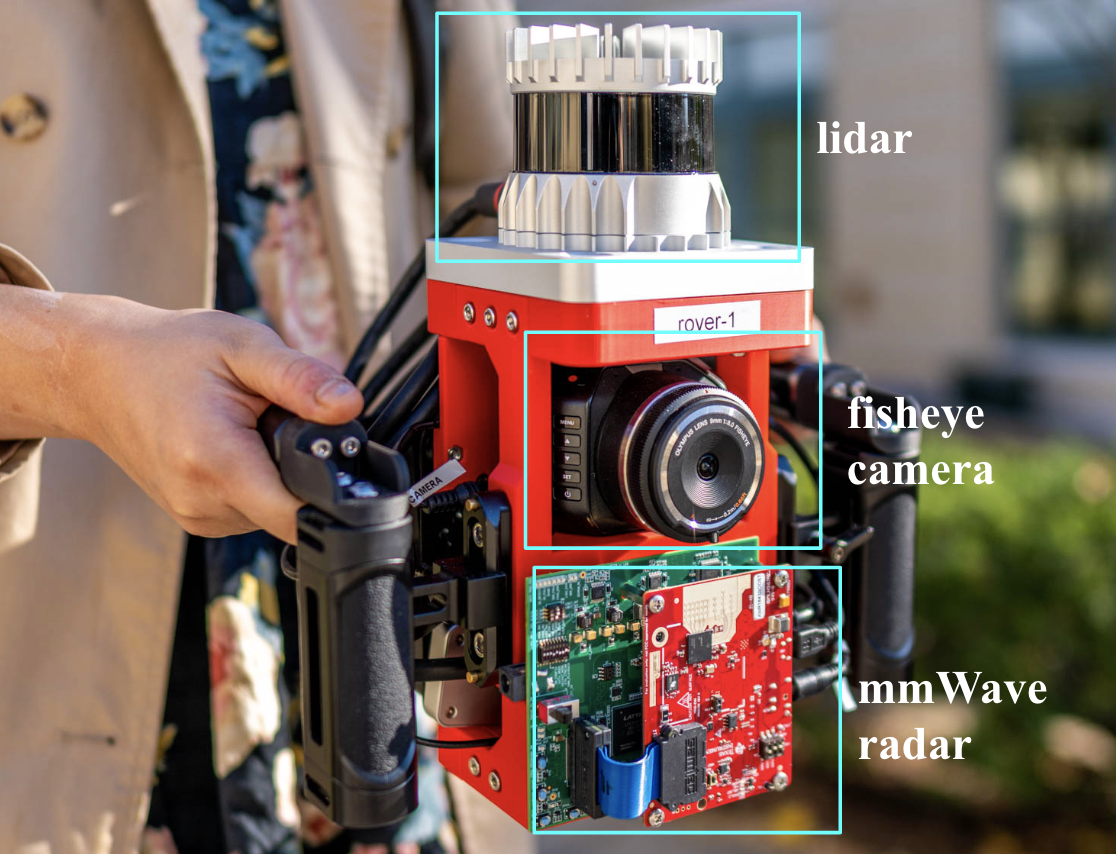}
\caption{
Image of our hand-held data collection rig with three key components labeled.
}
\label{onnerf}
\end{figure}

\subsection{Data capture rig and pose processing}
We build a hand-held rig for data collection with time-synced mmWave radar, fisheye camera, and lidar collecting data at 30 fps, 30 fps and 10 fps respectively.

For each sequence, we run COLMAP to get camera poses $\bm{A_c}$ (can be broken down into rotation $\bm{R_c}$ and position $\bm{x_c}$ ). Since coordinate systems of camera and radar are calibrated, we can easily convert camera pose to radar pose $\bm{A_r}$ (or $\bm{R_r}$ and $\bm{x_r}$) by interpolating into the sequence of camera poses with synced-timestamps followed with a transformation. We use the proposed scale estimation approach mentioned in Section 3.5 of the main paper to obtain scale of the scene for estimating ego-velocity required for our system.
\subsection{Multimodal pose refinement }
While COLMAP provides fairly accurate poses , radar ray tracing requires accurate \textbf{velocity}, so we design a pose and velocity refinement module by learning a per-frame pose and velocity offset $\Delta\bm{x_c}, \Delta\bm{R_c}, \Delta\bm{v_c}$ and regularized through:
\begin{enumerate}
    \item L2 regularization on the offsets: $L_{regp} = ||\Delta\bm{x_c}||_2^2 + ||\Delta\bm{R_c}||_2^2 +
    ||\Delta\bm{v_c}||_2^2$

     \item L2 loss that enforces velocity to be close to derivative of position $L_{regv} = ||d(\bm{x_c}) / dt - (\bm{v_c}+\Delta\bm{v_c})||_2^2$

     \item L2 regularization on acceleration 
     $L_{rega} = ||d(\Delta\bm{v_c}+\bm{v_c}) / dt||_2^2$

     \item kinematic loss: $L_{regk} = ||d_{window}(\bm{x_c}+\Delta\bm{x_c}) - d_{window}(\int \Delta\bm{v_c}+\bm{v_c}dt)||_2^2$

\end{enumerate}

  Note that it is possible to derive optimized velocity from optimized positions ($\bm{x}+\Delta\bm{x}$),  but we empirically found it to be more stable to keep a different  set of velocity offset and position offset and have them loosely connected through regularization. Such pose optimization scheme is shared across camera and radar as radar pose and velocity can be interpolated from camera poses and velocities. We show in Table \ref{tab:comparison} that our proposed pose refinement method improves on original DART\cite{huang2024dart} without pose refinement. 
\subsection{Evaluation traces break down}
\begin{figure*}[h]
    \centering
\includegraphics[width=\textwidth]{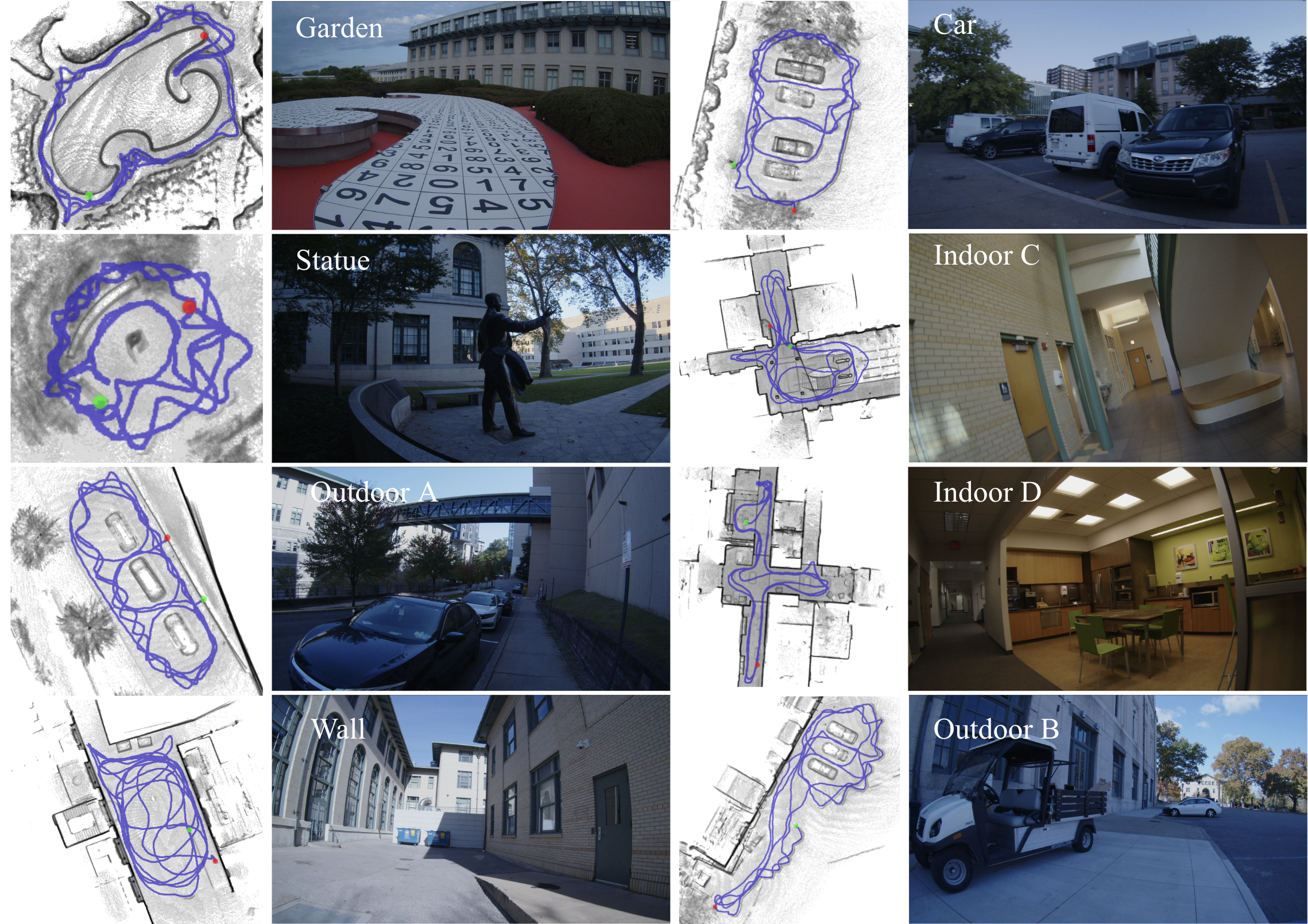}
\vspace{-1.5em}
\caption{
Visualization of the 8 indoor and outdoor scenes we collect and evaluate on in our experiments, in lidar map \textbf{(left)} and a sample RGB image \textbf{(right)}. 
}
\label{fig:scenes}
\end{figure*}

We show lidar map of the scene, trajectory and a RGB view of our 8 evaluation traces in Figure \ref{fig:scenes}. Number of images and radar frames range from 2000 to 3000. Image size used for training is 960x540 px.

\subsection{Existing multimodal camera-radar dataset}
We include a summary of existing multimodal datasets that contain raw radar measurement in Table  \ref{tab:other_datasets} in the main paper and also attached here that we further elaborate on. There is a lack of multimodal radar-camera dataset that contains raw single-chip radar measurement for scene reconstruction, which requires overlapping, and view-direction varying scene content across multiple measurements. Automotive datasets, such as RADIal \cite{Rebut_2022_CVPR} and RaDICaL~\cite{9361086}, offer limited variation in viewpoints and sensor height, making it difficult to use for evaluating novel view synthesis. RADIal \cite{Rebut_2022_CVPR} also uses Cascaded Radar with a complex modulation function, which makes modelling radar rendering equation difficult for inverse rendering. Coloradar \cite{kramer2022coloradar} does capture indoor senes that can be used for reconstruction, but uses high-resolution radar which is also beyond our sin-radar focus. Most public RGB-Radar datasets, such as NuScenes \cite{fong2021panoptic}, provide only {\em post-processed} CFAR data---not the {\em raw} measurements that we use. Hence we collect our own multimodal dataset and evaluate performance of our approach and baselines.

\begin{table*}[!h]
\centering
\footnotesize
\begin{tabular}{lllll}
\toprule
Dataset & Radar Type & Raw Data & View-direction Varying & Other Sensors \\
\midrule
RadarSim (Ours) & \textbf{Low Res} & \textbf{Yes} & \textbf{Yes} & Lidar, Camera, IMU \\
RADDet \cite{zhang2021raddet} & \textbf{Low Res} & \textbf{Yes} & No  & Camera \\
RADIal \cite{rebut2022raw} & High Res & \textbf{Yes} & No & Lidar, Camera, GPS \\
K-radar \cite{paek2022k} & High Res & \textbf{Yes} & No & Lidar, Camera, IMU, GPS \\
Coloradar \cite{kramer2022coloradar} & High Res & \textbf{Yes} & \textbf{Yes} & Lidar, IMU \\
RaDICal \cite{lim2021radical} & \textbf{Low Res} & \textbf{Yes} & No & Depth Camera, IMU \\

\bottomrule
\end{tabular}

\caption{\textbf{Comparison with other mmWave radar datasets with raw data.} We capture a dataset using a low-resolution single-chip radar and cover scene content from multiple views directions and positions.} 
\label{tab:other_datasets}
\end{table*}

 \section{Method details}
 \begin{figure*}[h!]
    \centering
\includegraphics[width=\textwidth]{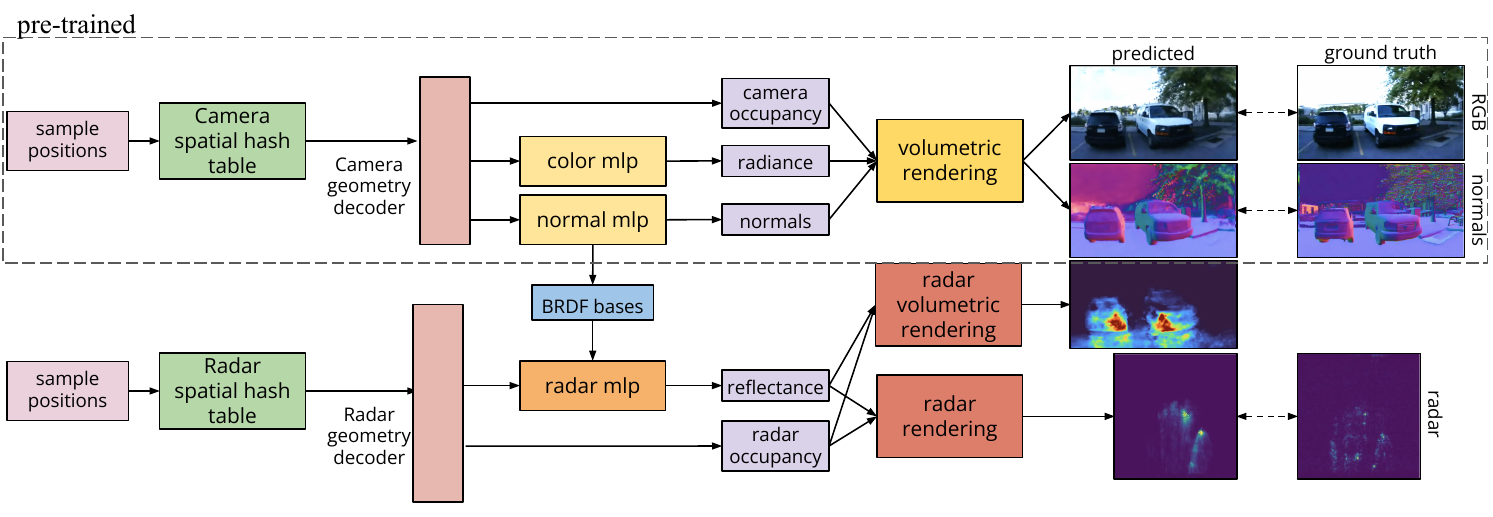}
\caption{
Architecture diagram of our framework.
}
\label{fig:archdia}
\end{figure*} 

 We elaborate here on technical details omitted in the main paper. See figure \ref{fig:archdia} for detailed architecture diagram.

  \subsection{Modelling radar view dependence with BRDF bases}
View dependence of Radar reflectance can be  broken into 2 scenarios, surface normal dependent reflectance and surface normal independent reflectance. The latter includes reflectance from retro-reflectance structures such as corners, bottom of car etc. ( where rays always bounce back in a particular direction) and inter-reflections. To model these 2 scenarios, we input to Radar MLP with BRDF bases to model normal dependent view dependence, and spherical harmonics encoded view directions to model normal independent view dependence. Geometry code is also input to the Radar MLP as both types of reflections are spatially varying. 

\subsection{Training details}
  Following DART \cite{huang2024dart}, we process radar raw data into Range-Doppler-Azmith frames with dimension size 128, 128, 8. At each training iteration, a batch of radar Doppler columns $Y_{r}$  are sampled to form radar rays.  Radar rays are sampled on a cone with directions determined by  velocity of the sensor at the current frame, and apex angle determined by the dot product between speed and Doppler value of the sampled column.

   \paragraph{Radar rendering.} Radar ray batch is sampled using a fine-tuned radar proposal sampler which outputs different sampling weights from camera. The hash table and density MLP corresponding to $f_{geo_r}$ are initialized with pre-trained $f_{geo_c}$ and optimized using radar measurements. They are queried to obtain radar occupancy and a geometry code, which is decoded and fed to the Radar MLP along with SH encoded view directions and  our proposed BRDF bases to decode into radar reflectance. Reflectance and radar occupancy are assigned to each range bin as discussed in Sec 3.3 in the main paper, and rendered with DART rendering equation to synthesize the input Doppler column $\hat{Y_r}$  \cite{huang2024dart} .

 \subsection{Loss Functions}
 RGB model is supervised with losses used in Nerfacto. Our model is supervised with $L_1$ reconstruction loss and SSIM loss for radar measurement, binary cross entropy loss between radar and camera occupancy to constrain radar geometry to be close to camera geometry, as well as auxiliary losses for pose regularization (mentioned in Section 2.2), proposal sampling network and normals. We list the loss functions here:
   \paragraph{Reconstruction loss}
   \begin{equation}
       L_{r} = ||Y_r - \hat{Y_r}||_1
          \end{equation}

     \paragraph{SSIM loss}
   \begin{equation}
       L_{ssim} = 1 - SSIM(Y_r, \hat{Y_r})
          \end{equation}
   \paragraph{Geometry consistent loss}
     \begin{equation}
    L_{bce} = -\left[ \alpha_{c} \log(\alpha_{r}) + (1 - \alpha_{c}) \log(1 - \alpha_{r}) \right]
          \end{equation}
    \paragraph{Interlevel Losses} 
    We fine-tune the proposal sampler for radar with a proposal loss $L_{prop}(\bm{t}, \bm{w}, \hat{\bm{t}}, \hat{\bm{w}})$ discussed in \cite{barron2022mipnerf360} to encourage histgram of rendering weights $\hat{\bm{w}}$ queried from the proposal network at samples $\hat{\bm{t}}$ to match the rendering weight $\bm{w}$ of the geometry field at a set of different sample positions $\bm{t}$. 
    Specifically, 
     \begin{equation}
       L_{prop}(\bm{t}, \bm{w}, \hat{\bm{t}}, \hat{\bm{w}}) = 
       \Sigma_i \frac{1}{\bm{w}_i} max(0, \bm{w_i}-
       bound(\hat{\bm{t}}, \hat{\bm{w}}, T_i))
   \end{equation}
   where $ bound(\hat{\bm{t}}, \hat{\bm{w}}, T_i)$ is the sum of proposal weights $\hat{\bm{w}}$ in interval $i$. This loss function penalizes the proposal weights that under-estimates the rendering weight distribution from geometry field. 
   Please refer to  \cite{barron2022mipnerf360} for details about this loss function. 
   We apply this loss to radar as $L_{prop_r}$, to enforce the proposal network to generate sampling weights that focus on radar geometry. Radar rendering weights are computed by 
    \begin{align}
    \bm{w_r} =\alpha_r(\bm{t}_i)
    \prod_{j < i}( 1-\alpha_r(\bm{t}_j))
   \end{align} 
    \paragraph{Normal Supervision Losses}

    We supervise our normal prediction MLP with psudo ground truth normal $\bm{n_{gt}}$ from a monocular normal estimator \cite{fu2024geowizard} by first converting it to world coordinate using camera extrinsics $\bm{A_c}$:

    \begin{align}
        L_{norm} = ||\bm{n}- \bm{A_c}\bm{n_{gt}} ||_2^2
    \end{align}

    Our combined loss is 

    \begin{multline}
    L = \lambda_{r}L_{r}  + \lambda_{bce}L_{bce} +  \lambda_{ssim}L_{ssim} + \lambda_{prop_r}L_{prop_r}+   \\
    \lambda_{norm}L_{norm} + \lambda_{norm_g}L_{norm_g} + \lambda_{norm_o}L_{norm_o} +  \lambda_{regp}L_{regp} \\
    + \lambda_{regv}L_{regv} +\lambda_{rega}L_{rega}+
    \lambda_{regk}L_{regk}
    \end{multline}

    In this equation, $L_{norm_g} + L_{norm_o}$ are adapted from \cite{verbin2022ref} to guide the predicted normal with gradient direction of the camera density field, and encourage normals to point outward from a surface. We use $\lambda_{r}=1e^{-3}$ for outdoor scenes and $\lambda_{r}=1e^{-4}$ for indoor scenes where abundance of multi-path reflections result in overall high reflectance in the scenes. We choose $\lambda_{bce}=0.01$ and $\lambda_{ssim}=0.01$.  We choose $\lambda_{norm}=0.1$ to obtain strong supervision from ground truth normal. We follow default values in Nerfstudio and use $\lambda_{prop_r}=1$, $\lambda_{norm_g}=1e^{-3}$, 
    $\lambda_{norm_o}=1e^{-4}$. For pose refinement, 
    we use $\lambda_{regp}=1e^{-3}$, $\lambda_{regv}=1$, 
    $\lambda_{rega}=5e^{-3}$, $\lambda_{regk}=1$.

  \subsection{Implementation Details}
 We implement our pipeline using Pytorch in Nerfstudio \cite{nerfstudio} based on Nerfacto-big \cite{nerfstudio} . For training we used a single 24 GB Nvidia Rtx3090 GPU. Each sequence is trained on RGB images first for 50k iterations and fine-tuned on radar data for 30k iterations. Training time is around 2 hours. Learning rate for the radar model is $1e^{-2}$ annealed to $1e^{-4}$ after 30k steps, and for pose refinement is $1e^{-3}$ annealed to $1e^{-4}$ after 5k steps. We list model hyperparameters in Table \ref{tab:params}.

\begin{table*}[!h]
\centering
\footnotesize
\begin{tabular}{lll}
\toprule
\textbf{Model}                                                                        & \textbf{Configuration}                & \textbf{Value}                                        \\ \hline
SH                                                                           & degree                       & 25                                           \\ \hline
BRDF bases                                                                   & number                       & 11                                           \\ \hline
\begin{tabular}[c]{@{}l@{}}Proposal\\ Hash\\ Encoding level 0,1\end{tabular} & \# of levels                 & 5, 5                                         \\
                                                                             & Hash table size              & 2\textasciicircum{}17, 2\textasciicircum{}17 \\
                                                                             & \# of feature dim. per entry & 2,2                                          \\
                                                                             & Coarse resolution            & 16,16                                        \\
                                                                             & Fine resolution              & 128, 256                                     \\
                                                                             & Decoder feature dim          & 16,16                                        \\
                                                                             & Number of layer              & 2,2                                          \\
                                                                             & \# of ray samples            & 512, 256                                     \\ \hline
\begin{tabular}[c]{@{}l@{}}Hash\\ Encoding\end{tabular}                      & \# of levels                 & 16                                           \\
                                                                             & Hash table size              & 2\textasciicircum{}21                        \\
                                                                             & \# of feature dim. per entry & 2                                            \\
                                                                             & Coarse resolution            & 16                                           \\
                                                                             & Fine resolution              & 2048                                         \\
                                                                             & \# of ray samples            & 64                                           \\ \hline
\begin{tabular}[c]{@{}l@{}}Density\\ MLP\end{tabular}                        & \# of hidden layer           & 2                                            \\
                                                                             & \# of neuron per layer       & 128                                          \\
                                                                             & Output activation            & Exp                                          \\
                                                                             & Density feature dim          & 15                                           \\ \hline
Radar MLP                                                                    & \# of hidden layer           & 2                                            \\
                                                                             & \# of neuron per layer       & 128                                          \\
                                                                             & Output activation            & None                                         \\ \hline
Color MLP                                                                    & \# of hidden layer           & 2                                            \\
                                                                             & \# of neuron per layer       & 128                                          \\
                                                                             & Output activation            & Sigmoid                                      \\ \hline
Normal MLP                                                                   & \# of hidden layer           & 2                                            \\
                                                                             & \# of neuron per layer       & 64                                           \\
                                                                             & Output activation            & None                                        
\\
\bottomrule
\end{tabular}%
\caption{ List of hyperparameters used in our architecture.}
\label{tab:params}
\end{table*}

\section{Experiment details and additional result}
   \subsection{Evaluation metric and denoising procedure}
We calculate PSNR and SSIM values between ~\approach~ and ground truth for evaluation and comparison against baselines. Because most of the radar frame consists of noise, we design a denoising procedure by  finding the noise threshold for each dataset. The noise threshold is calculated by fitting a chi-square distribution to the empty Doppler columns  of  each dataset (where speed is smaller than the Doppler values) and take a p-value of 0.01 of the noise distribution as the noise threshold. During evaluation,  ground truth and synthesized range-Doppler frames are clipped at 0.01 and 99.99 percentile of the ground truth over the entire dataset and normalized to 0 and 1 to calculate SSIM and PSNR. Areas where the ground truth frames are  below this threshold are considered to be noise and ignored during both PSNR and SSIM calculation. 

 \subsection{Description of baselines}
 We run a pytorch version of DART \cite{huang2024dart} using code provided by the authors. The parameter for the hashtable and geometry decoder is set to match the size of \approach. We use the same set of poses for Radar for the baseline and our approach, where they are time-interpolated from COLMAP-derived camera poses. All comparisons included in the main paper with DART use additional pose refinement described in Section 2  for fair comparison with \approach. We include comparison with DART without pose refinement in Table \ref{tab:comparison}. It can be observed that our proposed pose refinement scheme improves DART's novel view synthesis quality as the model is less prone to overfitting to the inaccuracies in poses. We also implement RadarFields \cite{borts2024radar} as an additional radar-only baseline. Since this approach is designed for high resolution radar, it only uses range-azimuth. To compare with our method, we train it on range-azimuth slices  of the input data only and evaluate on range-doppler-azimuth synthesis quality. We also include result in Table \ref{tab:comparison}. Although this method includes pose refinement, its performance is inferior to that of DART with pose refinement, suggesting the importance of leveraging Doppler for low resolution radar.

\begin{table*}[!h]
\centering
\footnotesize
\resizebox{\textwidth}{!}{%
\begin{tabular}{ccccccccccccccccccc}
\toprule
 & \multicolumn{2}{c}{\textbf{Car}} & \multicolumn{2}{c}{\textbf{Garden}} & \multicolumn{2}{c}{\textbf{Statue}} & \multicolumn{2}{c}{\textbf{Wall}} & \multicolumn{2}{c}{\textbf{Outdoor A}} & \multicolumn{2}{c}{\textbf{Outdoor B}} & \multicolumn{2}{c}{\textbf{Indoor C}} & \multicolumn{2}{c}{\textbf{Indoor D}} & \multicolumn{2}{c}{\textbf{Aggregated}} \\ \midrule
 & SSIM & PSNR & SSIM & PSNR & SSIM & PSNR & SSIM & PSNR & SSIM & PSNR & SSIM & PSNR & SSIM & PSNR & SSIM & PSNR & SSIM & PSNR \\ \hline
CFAR & 0.694 & 24.9 & 0.415 & 24.1 & 0.711 & 17.9 & 0.772 & 27.1 & 0.502 & 20.2 & 0.734 & 25.3 & 0.774 & 28.0 & 0.770 & 28.2 & 0.671 & 24.5 \\
Lidar & 0.782 & 29.4 & 0.470 & 26.7 & 0.830 & 28.9 & 0.796 & 27.9 & 0.595 & 26.8 & 0.782 & 28.6 & 0.805 & 29.1 & 0.809 & 29.2 & 0.734 & 28.3 \\
Nearest Neighbor & 0.733 & 26.5 & 0.592 & 24.6 & 0.798 & 26.2 & 0.810 & 24.6 & 0.580 & 22.9 & 0.726 & 24.3 & 0.782 & 26.6 & 0.785 & 27.3 & 0.726 & 25.4 \\
RadarFields~\cite{borts2024radar} & 0.831 & 29.3 & 0.603 & 27.0 & 0.869 & 29.0 & 0.781 & 27.1 & 0.667 & 24.4 & 0.773 & 27.5 & 0.814 & 28.6 & 0.831 & 29.3 & 0.771 & 27.8 \\
DART ~\cite{huang2024dart} without pose refinement & 0.832 & 29.8 & 0.616 & 27.1 & 0.866 & 29.1 & 0.809 & 27.0 & 0.691 & 26.3 & 0.797 & 27.1 & 0.817 & 28.5 & 0.846 & 29.3 & 0.784 & 28.0 \\
DART~\cite{huang2024dart} with pose refinement & 0.850 & 30.5 & 0.658 & \textbf{27.7} & 0.887 & 30.1 & 0.818 & 27.1 & 0.702 & 26.6 & 0.801 & 27.5 & 0.827 & 28.9 & 0.853 & 29.5 & 0.799 & 28.5 \\ \midrule
\approach~ (ours) & \multicolumn{1}{r}{\textbf{0.859}} & \multicolumn{1}{r}{\textbf{30.8}} & \multicolumn{1}{r}{\textbf{0.669}} & \multicolumn{1}{r}{27.5} & \multicolumn{1}{r}{\textbf{0.889}} & \multicolumn{1}{r}{\textbf{30.2}} & \multicolumn{1}{r}{\textbf{0.851}} & \multicolumn{1}{r}{\textbf{28.3}} & \multicolumn{1}{r}{\textbf{0.741}} & \multicolumn{1}{r}{\textbf{27.6}} & \multicolumn{1}{r}{\textbf{0.853}} & \multicolumn{1}{r}{\textbf{29.1}} & \multicolumn{1}{r}{\textbf{0.845}} & \multicolumn{1}{r}{\textbf{29.5}} & \multicolumn{1}{r}{\textbf{0.863}} & \multicolumn{1}{r}{\textbf{29.7}} & \multicolumn{1}{r}{\textbf{0.821}} & \multicolumn{1}{r}{\textbf{29.1}} \\ 
\bottomrule
\end{tabular}%
}
\caption{ 
\textbf{Per-scene break down comparison with baselines RadarFields \cite{borts2024radar} and  DART\cite{huang2024dart} with and without pose refinement.}
\approach~ achieves significantly higher PSNR and SSIM on outdoor scenes (\textit{Car, Statue, Wall, Outdoor A Outdoor B}) compared to baselines, as well as  higher performance on indoor scenes (\textit{Indoor C, Indoor D}), though the gains are less pronounced. This is because the effectiveness of leveraging surface normals to reproduce input radar recordings is reduced in indoor environments due to the increased presence of multipath reflections from corners and clutter. For Outdoor scene \textit{Garden}, DART slightly outperforms \approach~ as this scene is dominated by inter-reflections between the metallic edge of the garden and ground as well as from bushes which diffusely reflects radar signal, making our normal-dependent model less effective.}
\label{tab:comparison}
\end{table*}

\subsection{ Additional application:  geometry improvement for textureless region} 

\begin{figure}[!h]
    \centering
\includegraphics[width=0.5\textwidth]{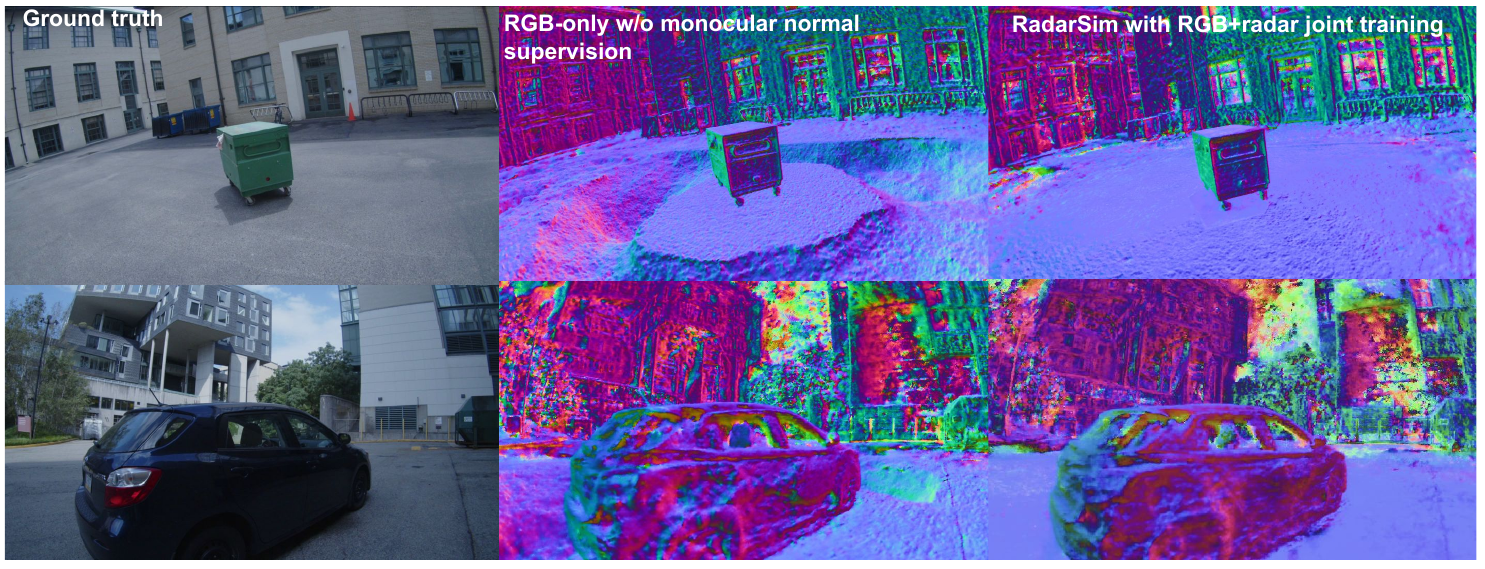}
\caption{Normal-dependent BRDF encoding allows \approach~ to improve reconstruction of textureless regions in a joint training setting, without using monocular-predicted normals.}
\label{fig:improve}
\end{figure}

We further illustrate the effectiveness of our BRDF encoding for modelling radar reflectance in a multimodal training framework where a single geometry field is optimized to represent radar and camera and trained simultaneous using radar and camera reconstruction losses. Without using monocular normal or depth priors, RGB-only reconstruction fails at identifying the true geometry for textureless regions. In a multimodal training setting, since our BRDF encoding effectively models radar view-dependence using predicted normals from the shared geometry, information from radar measurement can be propagated to the geometry and reconstruct textureless regions correctly as shown in figure \ref{fig:improve}. Comparison with other RGB-based or multimodal-based methods in improving geometry quality is left to future work.

\end{document}